\documentclass[final]{cvpr}

\usepackage{times}
\usepackage{epsfig}
\usepackage{graphicx}
\usepackage{amsmath}
\usepackage{amssymb}
\usepackage{multirow}
\usepackage{pifont}
\usepackage{booktabs}
\usepackage{epsfig}
\usepackage{booktabs,multirow}
\usepackage{graphics}
\usepackage{threeparttable}
\usepackage{color}
\usepackage[normalem]{ulem}
\usepackage{multirow}
\usepackage{float}
\usepackage{amsfonts}
\usepackage{bm}
\usepackage{enumitem}
\usepackage[numbers]{natbib}
\usepackage{array}
\usepackage[table]{xcolor}
\usepackage{colortbl}
\definecolor{myy}{RGB}{126,95,0}
\definecolor{mygray}{gray}{.9}
\definecolor{bblue}{RGB}{30,80,120}
\definecolor{mygray1}{gray}{.7}
\usepackage{bm}
\usepackage{mathtools}
\usepackage{subcaption}
\usepackage{siunitx}
\usepackage{amsfonts}
\usepackage{bm}
\usepackage{bbm}

\newcolumntype{I}{!{\vrule width 1pt}}

\definecolor{ggray}{RGB}{127,127,127}

\makeatletter
\newcommand{\thickhline}{%
	\noalign {\ifnum 0=`}\fi \hrule height 1pt
	\futurelet \reserved@a \@xhline
}
\makeatother

\usepackage{caption}
\usepackage[pagebackref=true,breaklinks=true,colorlinks,bookmarks=false]{hyperref}
\usepackage[utf8]{inputenc}

\usepackage{cleveref}
\crefname{section}{§}{§§}
\Crefname{section}{§}{§§}

\def\cvprPaperID{5988} % *** Enter the CVPR Paper ID here
\def\confYear{CVPR 2021}

\begin{document}

%%%%%%%%% TITLE
%\title{Robust Instance Discovery and Association for \\Unsupervised Video Object Segmentation}
\title{Target-Aware Object Discovery and Association for Unsupervised \\Video Multi-Object Segmentation}

\author{Tianfei Zhou$^{1}$,~~Jianwu Li$^{2}$\thanks{Corresponding author: \textit{Jianwu Li}.},~~Xueyi Li$^{2}$,~~Ling Shao$^{3}$  \\ 
$^1$ Computer Vision Laboratory, ETH Zurich, Switzerland \\ $^2$ School of Computer Science and Technology, Beijing Institute of Technology, China \\ $^3$ Inception Institute of Artificial Intelligence, UAE
	% \\
}

\maketitle
\thispagestyle{empty}

\begin{abstract}
	
	This paper addresses the task of unsupervised video multi-object  segmentation.  Current approaches follow a two-stage paradigm: 1) detect object proposals using pre-trained Mask R-CNN, and 2) conduct generic feature matching for temporal association using re-identification techniques. However, the generic features, widely used in both stages, are not reliable for characterizing unseen objects, leading to poor generalization. To address this, we introduce a novel approach for more accurate and efficient spatio-temporal segmentation. In particular, to address \textbf{instance discrimination}, we propose to combine foreground region estimation and instance grouping together in one network, and additionally introduce temporal guidance for segmenting each frame, enabling more accurate object discovery. For \textbf{temporal association}, we complement current video object segmentation architectures with a discriminative appearance model, capable of capturing more fine-grained target-specific information. Given object proposals from the instance discrimination network, three essential strategies are adopted to achieve accurate segmentation: 1) target-specific tracking using a memory-augmented appearance model; 2) target-agnostic verification to trace possible tracklets for the proposal; 3) adaptive memory updating using the verified segments. We evaluate the proposed approach on  DAVIS$_{17}$ and YouTube-VIS, and the results demonstrate that it outperforms state-of-the-art methods both in segmentation accuracy and inference speed.

\end{abstract}

\maketitle

\section{Introduction}

Unsupervised video object segmentation aims at automatically segmenting primary object(s) from the background in unconstrained videos, which is a fundamental vision task. This task has become increasingly popular due to its potential values in a wide range of real-world applications, \eg, video compression~\cite{hadizadeh2013saliency}, autonomous driving~\cite{geiger2012we}, and human-centric understanding~\cite{wang2021hierarchical,zhoucascaded,zhou2021differentiable}.  However, the task is  challenging due to the lack of prior knowledge about the target objects, as well as the challenging factors (\eg, occlusions, cluttered background, diverse motion patterns) carried by video data.

%due to the lack of prior knowledge about the target objects, it is very challenging to accurately discover and segment distinct objects, especially in complex scenarios. 

%Instance segmentation has been an active research topic in computer vision, with the goal to separate individual instances from observed scenes.  With the recent renaissance of deep neural networks and in particular the development of representative frameworks (\eg, Mask R-CNN~\cite{he2017mask}, FCOS~\cite{tian2019fcos}), instance segmentation has achieved tremendous progress in both accuracy and efficiency. However, most of current approaches were designed for static image parsing. How to extend the success of segmentation techniques to video-based applications (\eg, autonomous driving, robotics, and surveillance) remains a challenging problem.
%
%In this work, we focus on the task of automatically segmenting and tracking multiple objects in videos. Requiring no human interventions, the task can benefit numerous video-related applications  for more effecient video understanding. Compared to image-level instance segmentation, this task is extremely challenging 

Towards better segmenting the prominent foreground objects, early studies typically exploit saliency cues~\cite{wang2015saliency,faktor2014video} or objectness priors~\cite{lee2011key,zhang2013video,xiao2016track,zhou2016video,li2017learning} for identifying them. More recently, with the advent of deep neural networks, many learning-based models have been proposed to learn more discriminative video object patterns, by leveraging motion cues~\cite{tokmakov2017learningmotion}, addressing spatiotemporal features~\cite{jain2017fusionseg,zhou2020motion}, exploring multi-frame contextual information~\cite{lu2019see,wang2019zero,yang2019anchor} or using recurrent networks to capture sequential information~\cite{song2018pyramid}. Though impressive results have been achieved, these approaches mainly focus on foreground/background separation, hindering their applications in more practical multi-object scenarios.

Unsupervised video multi-object segmentation,
%\footnote{In this paper, we use the term `U-VOS' to indicate both single- and multi-object segmentation tasks. However, we claim that the proposed approach targets multi-object scenarios.}
 with an elegant and formal definition in~\cite{caelles20192019}, is more challenging as it requires not only discovering instance-agnostic, foreground  regions automatically, but also discriminating different object instances and associating the same identities over the entire sequence. To tackle this task, existing methods~\cite{luiten2020unovost,song2018pyramid,Wang2020PayingAT,zhou2020matnet} generally follow the conventional tracking-by-detection paradigm which performs in a top-down fashion to employ image-aware instance segmentation networks (\eg, Mask R-CNN~\cite{he2017mask}, SOLO~\cite{wang2019solo}) to detect object candidates in individual frames, and  associate them over consecutive frames based on object tracking or proposal re-identification (ReID). In addition, to avoid the negative impact of background objects, many studies~\cite{song2018pyramid,Wang2020PayingAT,zhou2020matnet} also rely on a foreground/background separation step to remove background proposals. Even though these approaches demonstrate compelling performance, they still suffer several limitations.~\textbf{1)} Directly using image-level instance segmentation networks is insufficient since they are trained on static images, neglecting the informative temporal context in videos.~\textbf{2)} Instance segmentation and foreground estimation are often separately considered by different networks, incurring high computational expense.~\textbf{3)} ReID-based matching networks, trained completely offline, focus more on general object appearance, while rarely capturing distinctive fine-grained features of specific targets.

In this work, we propose a novel approach for unsupervised multi-object segmentation in unconstrained videos. To address points \textbf{1)} and \textbf{2)}, we introduce an instance discrimination network (D-Net) for video object proposal.  The network performs in a bottom-up fashion and takes video temporal information into account to achieve better segmentation accuracy and efficiency. In particular, the D-Net includes two branches: a foreground estimation branch follows the typical design of fully convolutional networks to segment attention-grabbing objects, and an instance segmentation branch learns to predict the instance center as well as the offset from each pixel to its corresponding center for instance grouping. Rather than processing each frame independently, we consider segmentation of previous frames as an important guidance for segmenting the current frame. In this way, we integrate instance-agnostic and instance-aware segmentation together into one network for discovering temporal coherent object proposal.

To address point \textbf{3)}, we design a {target-aware tracking network (T-Net)} for associating object proposals of the same identities over each image sequence. Different from previous ReID-based matching techniques, we aim to learn target-specific appearance features for more robust object association. More specifically, the T-Net learns a discriminative appearance model for each object instance during the inference stage to predict a coarse but robust segmentation score of the target object. Note that the appearance model is more prone to drifting due to the lack of ground-truth annotations. Therefore, we further propose a target-agnostic backward verification module to examine the tracking results. The verified results are used as new training samples to update the appearance model online.

With above efforts, our algorithm achieves state-of-the-art results on the DAVIS$_{17}$ benchmark for video multi-object segmentation. It also  demonstrates compelling performance for video instance segmentation on YouTube-VIS. In addition, our approach obtains a better trade-off between segmentation accuracy and inference efficiency, running at about 10 FPS on images with 480p resolution.

To sum up, the contributions of the proposed approach are three-fold: 
\textbf{First}, we propose a novel bottom-up instance discrimination network which takes advantage of temporal context information in videos for more accurate segmentation. The network couples foreground discovery and instance grouping together, benefiting from multi-tasking and improving the inference efficiency. \textbf{Second}, we introduce a target-aware tracking model for online matching of object proposals. Compared with target-agnostic approaches, our method can better capture the appearance information of the target objects, yielding more robust association. \textbf{Third}, our approach achieves compelling performance on the popular DAVIS$_{17}$ and YouTube-VIS benchmarks. Furthermore, its high inference speed enables our method to support a wide variety of practical applications.

\begin{figure*}[t]
	\begin{center}
		\includegraphics[width=\linewidth]{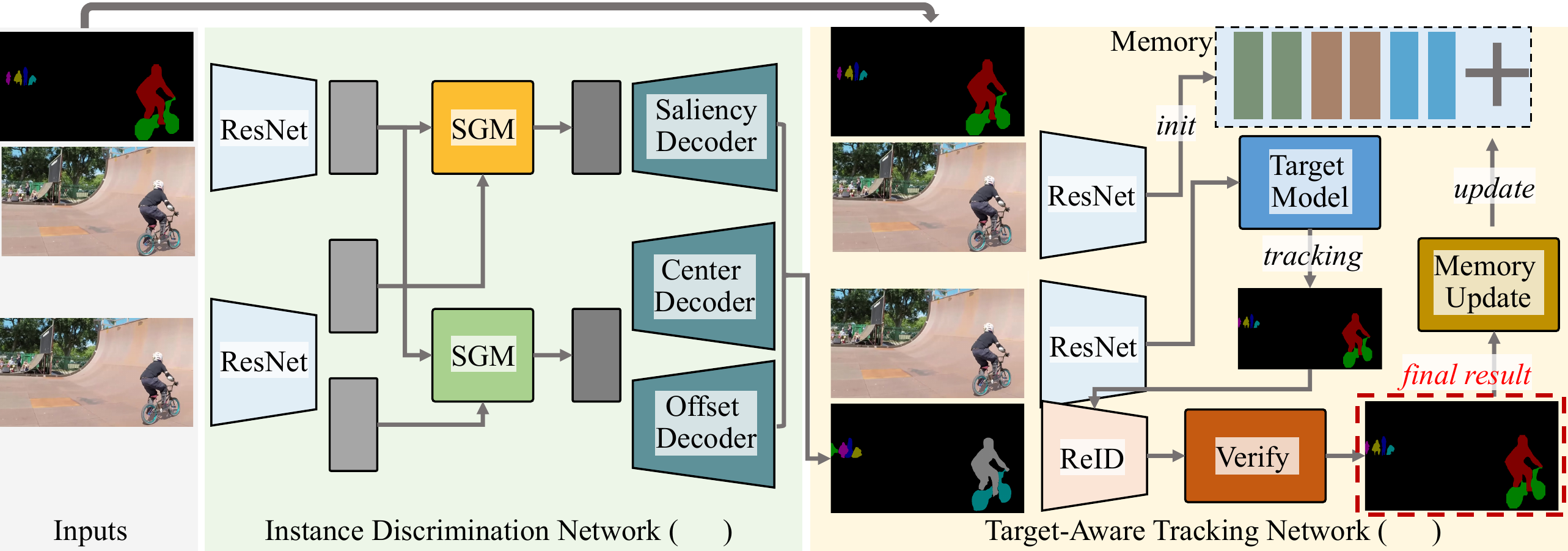}
				\put(-282,3){\small \S\ref{sec:dnet}}
				\put(-58,3){\small \S\ref{sec:tnet}}
				\put(-485,85){\small $I_{t-1}\& S_{t-1}$}
				\put(-470,30){\small $I_t$}
				\put(-392,151){\small $\bm{X}_{t-1}^{\text{SGM}}$}
				\put(-392,101){\small $\bm{X}_{t}^{\text{SAL}}$}
				\put(-392,57){\small $\bm{X}_{t}^{\text{INS}}$}
				\put(-316,151){\small $\hat{\bm{X}}_{t}^{\text{SAL}}$}
				\put(-316,79){\small $\hat{\bm{X}}_{t}^{\text{INS}}$}
	\end{center}
	\vspace{-16pt}
	\captionsetup{font=small}
	\caption{\textbf{Detailed illustration of our approach} for unsupervised video multi-object segmentation. }

	\label{fig:pipeline}
	\vspace{-10pt}
\end{figure*}
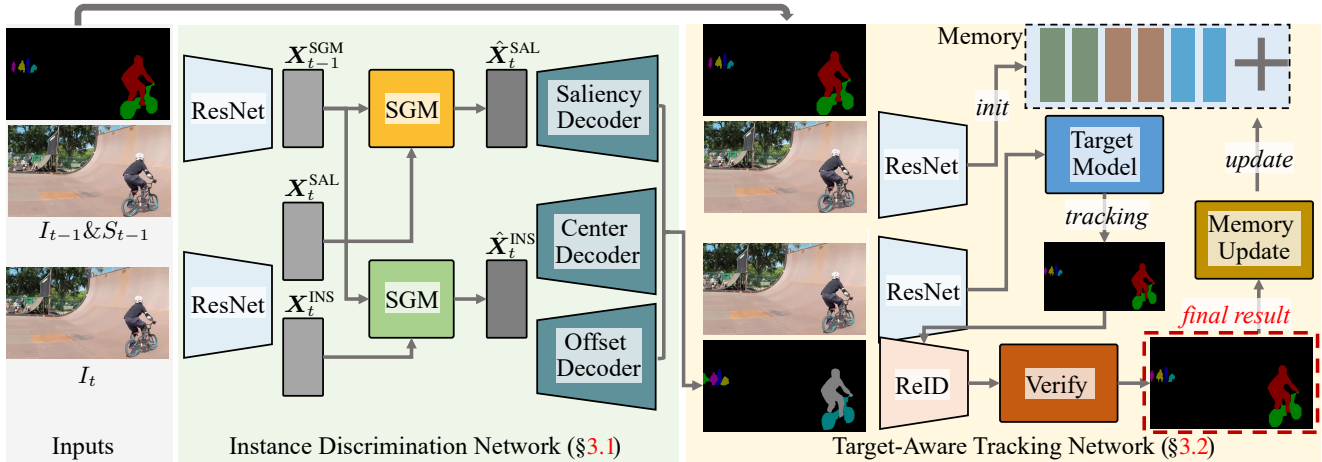

\section{Related Work}

\noindent\textbf{Unsupervised Video Object Segmentation.}
The task of U-VOS aims to segment conspicuous and eye-catching objects without any human intervention. Most current research efforts focus on segmenting the prominent foreground objects in unconstrained videos. Earlier methods typically rely on motion analysis, \eg, extracting motion information from sequential images to understand object movement. For example, a large number of works\!~\cite{brox2010object,fragkiadaki2012video,ochs2013segmentation,faisal2019exploiting} employ sparse point trajectories to capture long-term motion information and segment the objects which are moving significantly in relation to the background. However, these methods are not robust because they rely heavily on optical flow estimation and feature matching, and thus may easily fail in the presence of occlusions, fast motion or appearance changes.
%One major limitation of these methods is that they are constrained to the accuracy of optical flow estimation, thus easily fail in the presence of highly non-rigid motions. 
To address this limitation, later methods employ saliency cues\!~\cite{faktor2014video,wang2015saliency} and object proposal ranking\!~\cite{lee2011key,zhang2013video,perazzi2015fully} to better identify the main objects. These \textit{non-learning} methods are confined by the limited representative ability of handcrafted features. 

More recently, with the renaissance of artificial neural networks in computer vision, \textit{deep learning} based solutions are now dominant in this field. For example, AGS\!~\cite{Wang2020PayingAT} proposes a dynamic visual attention-driven model for video object segmentation, while\!~\cite{wang2019zero,lu2019see} mine higher-order relations between video frames, resulting in more comprehensive understanding of video content and more accurate foreground estimation. Moreover, many approaches discover the motion patterns of objects\!~\cite{tokmakov2017learningmotion} as complementary cues to object appearance. This is typically achieved within two-stream networks\!~\cite{jain2017fusionseg,tokmakov2017learning,faisal2019exploiting,hu2018unsupervised,zhou2020motion,Li_2019_ICCV}, in which an RGB image and the corresponding flow field are separately processed by two independent networks and the results are fused to produce the final segmentation. To avoid the expensive computation of optical flow, some methods\!~\cite{wang2017video,lu2020video} directly feed consecutive frames into the networks and automatically learn spatiotemporal feature representations.

%U-VOS aims to segment conspicuous and eye-catching objects without any human intervention. Most current research efforts focus on segmenting all primary objects together, leading to category-agnostic solutions. Among them, \textit{non-deep learning} approaches are based on hand-crafted features (\eg, color, texture) and rely on certain heuristics (\eg, object saliency\!~\cite{faktor2014video,wang2015saliency,wang2017saliency}, object proposal ranking\!~\cite{lee2011key,zhang2013video,perazzi2015fully}, trajectory clustering\!~\cite{brox2010object,ochs2011object,fragkiadaki2012video,ochs2013segmentation,keuper2015motion}). Recently, with the renaissance of neural networks in computer vision, \textit{deep learning} based solutions are now dominant in this field. For example,\!~\cite{Wang_2019_CVPR} proposes a dynamic visual attention-driven model for video object segmentation, and\!~\cite{wang2019zero,lu2019see} mine higher-order relations between video frames, resulting in more comprehensive understanding of video content and more accurate foreground estimation. Moreover, many approaches discover the motion patterns of objects\!~\cite{tokmakov2017learningmotion} as complementary cues to object appearance. This is typically achieved within two-stream networks\!~\cite{jain2017fusionseg,tokmakov2017learning,faisal2019exploiting,hu2018unsupervised,zhou2020motion}, in which an RGB image and the corresponding flow field are separately processed by two independent networks and the results are fused to produce the final segmentation.

\noindent\textbf{Unsupervised Video Multi-Object Segmentation.}
Unlike the aforementioned U-VOS approaches that pay more attention to learning powerful object representations for foreground object discovery, in the multi-object setting, the challenges become how to discover and segment each object that captures human attention, and how to associate the objects through the whole sequence. RVOS\!~\cite{ventura2019rvos} delivers an end-to-end recurrent neural network, in which the spatial recurrence helps to discover object instances in each frame, while the temporal recurrence finds the matching between instances in different frames. This work represents an early attempt
towards end-to-end learning for unsupervised video multi-object segmentation. However, RVOS is weak in the segmentation performance due to its limited capability in instance discrimination. Current trends\!~\cite{luiten2020unovost,zhou2020matnet} follow a two-stage pipeline, in which object proposals are first discovered using Mask R-CNN\!~\cite{he2017mask}, and the association is conducted using greedy- or ReID-based matching techniques.

%\noindent\textbf{Unsupervised Video Object Segmentation (UVOS).} UVOS aims to segment conspicuous video objects without any test-time human intervention. Most current research efforts focused on segmenting all primary objects together. These methods avoid the dilemma of data association, and pay more attention to enrich object representations for automatic object discovery. Specifically, they learned motion patterns to separate independent objects and camera motion\!~\cite{tokmakov2017learningmotionpattern}, mined high-order contextual relationships among video frames\!~\cite{lu2019see,wang2019zero,yang2019anchor}, or  exploited two-stream neural networks\!~\cite{dutt2017fusionseg,zhou2020motion}. However, in the instance-level multi-object setting, the main challenge becomes how to associate different objects across frames. Recent leading approaches\!~\cite{li2018video,luiten2020unovost} solved this by feature matching based ReID. Though impressive, they suffer from the limited representability of generic features in characterizing specific objects, which poses great difficulties for distinguishing similar objects. In contrast, we propose to learn target-specific features for robust instance tracking, and introduce a global matching strategy to improve the tracking results.

\noindent\textbf{Instance Segmentation in Images and Videos.}
In recent years, image-level instance segmentation has attracted great research interests, which extends semantic segmentation~\cite{long2015fully,li2020group,wang2021exploring,yao2021weakly} to assign different labels for separate instances of objects belonging to the same class. Driven by the success of R-CNN\!~\cite{girshick2015fast}, current dominant instance segmentation methods follow a detect-then-segment framework. Earlier methods\!~\cite{pinheiro2015learning,dai2016instance} learn to propose segment candidates, and then classify them by Fast R-CNN\!~\cite{girshick2015fast}. These methods conduct segmentation before recognition, which are slow and less accurate. 
%FCIS\!~\cite{li2017fully} combines proposal segmentation and object detection to predict a set of position sensitive output channels in a fully convolutional manner. These channels simultaneously address the predictions of object categories, bounding boxes, and segmentation masks, making the system fast. 
Mask R-CNN\!~\cite{he2017mask} introduces an extra ROI segmentation head into Faster R-CNN\!~\cite{ren2015faster} and a new assignment operator, \ie, ROIAlign, to better align the ROI features with inputs. Along this line, some works improve the performance by employing cascade inference\!~\cite{cai2019cascade}, low-level feature enhancement\!~\cite{liu2018path}, and multi-tasking\!~\cite{chen2019hybrid}. However, for complicated scenarios with many instances, the inference time of two-stage methods is unacceptably high, since it is proportional to the number of instances. The resolution of ROI features and resulting masks are coarse, resulting in poor segmentation of object boundaries. 

To cope with these drawbacks, many recent works favor bottom-up instance segmentation. These approaches are often box-free and thus not restricted by anchor locations and scales, naturally benefiting from the inherent advantages of fully convolutional networks. For example,~\cite{de2017semantic,gao2019ssap} learn discriminative embeddings to group the pixels into an arbitrary number of object instances. SOLO\!~\cite{wang2019solo} introduces a direct instance segmentation method that can predict instance segmentation in one shot without additional grouping post-processing. AdaptIS\!~\cite{sofiiuk2019adaptis} first generates point proposals as representations of instances, and then sequentially predicts the corresponding segmentation mask for each detected proposal. PolarMask\!~\cite{xie2020polarmask} utilizes the polar representation to encode masks and transforms per-pixel mask prediction to distance regression.

Though these methods only focus on image-level segmentation, we emphasize that they have motivated a number of video analysis tasks, such as video object/instance segmentation~\cite{yang2019video,hu2017maskrnn,2020Classifying,DAVIS2020-Unsupervised-2nd} and multi-object tracking~\cite{zhou2020tracking,voigtlaender2019mots}. In this work, we further propose a novel bottom-up approach for segment proposal generation in videos. Instead of frame-by-frame segmentation, we take advantage of the segmentation results in previous frames as temporal guidance, yielding more robust results.

%\subsection{Discrimintive Appearance Models}\label{sec:II-C}  
%Appearance models have been widely explored in online visual tracking\!~\cite{danelljan2017eco,robinson2020learning} to capture target object appearance.  Some recent efforts discriminatively learn convolution filters using efficient optimization (\eg,  Conjugate Gradient\!~\cite{danelljan2017eco}, Gauss-Newton\!~\cite{robinson2020learning}) to distinguish target from background. In this work, with a similar spirit of\!~\cite{robinson2020learning}, we build a target-specific appearance model and adapt it into our instance-level unsupervised video object segmentation scenario.

%\noindent\textbf{Video Instance Segmentation.}

\section{Methodology}\label{section:3}

%\subsection{Overview}

Given a video $\mathcal{I}\!=\!\{I_t\}_{t=1}^N$ with $N$ frames $I_t\!\in\!\mathbb{R}^{3\times h \times w}$ with spatial size $h\times w$, the goal of unsupervised video multi-object segmentation is to automatically generate a collection of non-overlapping segment tracks, each for an individual instance. As shown in~\figref{fig:pipeline}, we decompose the problem into two sub-tasks: 1) discover object instances using the D-Net (\S\ref{sec:dnet}) and 2) associate all instances of the same identity over the entire sequence with the T-Net (\S\ref{sec:tnet}).

%To achieve this, we tackle the task through an efficient two-stage framework, wherein the D-Net is devised for video instance discovery, while the T-Net sequentially associates the instances across frames by learning target-specific appearance cues. Next, we describe our approach in detail.

%Given a video sequence $\mathcal{I}$, the goal of unsupervised segmentation is to automatically generate a collection of non-overlapping segment tracks. We propose a two-step approach to achieve this: \textit{object instance discovery} and \textit{target-aware adaptive tracking}. In the first step, we devise a bottom-up (\ie, proposal-free) instance segmentation network to obtain a set of candidate object instances. These instances are then sequentially tracked using a target-aware adaptive tracking algorithm. Next, we describe our approach in detail.

\subsection{Instance Discrimination Network (D-Net)}\label{sec:dnet}

The D-Net consists of four major components: 1) a backbone network for feature extraction; 2) a segmentation guidance module to employ previous segmented masks to enrich the feature representations; 3) a foreground estimation head for primary object prediction; and 4) an instance segmentation head for instance-level prediction.

\noindent\textbf{Feature Extraction.}
Given the video frame $I_t$ at time $t$, we use a backbone CNN model to extract convolutional features $\bm{X}_t^{\text{SAL}}\!\in\!\mathbb{R}^{W\times H \times C}$, $\bm{X}_t^{\text{INS}}\!\in\!\mathbb{R}^{W\times H \times C}$, where $W$, $H$ and $C$ represent the width, height and channel number of the 3D tensors, respectively. $\bm{X}_t^{\text{SAL}}$ and $\bm{X}_t^{\text{INS}}$ indicate two task-specific features that are responsible for salient foreground estimation and instance-aware segmentation, respectively. To achieve this, we take the convolutional blocks of ResNet-50\!~\cite{he2016deep} as the backbone, and modify the last residual block with an atrous convolution with rate $2$ to enlarge the receptive field. Furthermore, in order to extract task-specific features, we augment the backbone network with two parallel atrous spatial pyramid pooling (ASPP) modules\!~\cite{chen2017deeplab}. ASPP applies several parallel atrous convolutions with different rates to further increase the receptive field. In our model, we design each ASPP to have 1) one $1\times1$ convolution and three $3\times3$ convolutions with astrous rates $=(6,12,18)$ and 2) a global average pooling layer on the last feature map of the backbone to obtain global context information. The resulting features from all branches are then concatenated and passed through an extra $1\times1$ convolutional layer to obtain  $\bm{X}_t^{\text{SAL}}$ and $\bm{X}_t^{\text{INS}}$. Both of these tensors have $C\!=\!256$ channels and output strides of 16.

\noindent\textbf{Segmentation Guidance Module (SGM).}
Rather than directly using the image-level feature representations (\ie, $\bm{X}_t^{\text{SAL}}$ and $\bm{X}_t^{\text{INS}}$) for segmentation prediction, we propose to exploit the inherent correlation among video frames for better results. Particularly, the segmentation mask $S_{t-1} $of the previous frame $I_{t-1}$ is leveraged as a guidance to improve the representation in the current frame. We introduce an extra convolutional branch whose input is the concatenation of $I_{t-1}$ and $S_{t-1}$. The input is processed with a similar backbone and ASPP module to obtain feature $\bm{X}_{t-1}^{\text{SGM}}$. Then, two segmentation guidance modules  are used to enrich the feature representations as follows:
\begin{align}\small
\hat{\bm{X}}{}_t^{\text{SAL}} &= \mathcal{F}^{\text{SGM}}(\bm{X}_t^{\text{SAL}}, \bm{X}_{t-1}^{\text{SGM}}), \\
\hat{\bm{X}}{}_t^{\text{INS}} &= \mathcal{F}^{\text{SGM}}(\bm{X}_t^{\text{INS}}, \bm{X}_{t-1}^{\text{SGM}}).
\end{align}
Each guidance module $\mathcal{F}^{\text{SGM}}$ has a squeeze-and-excitation structure~\cite{hu2018squeeze}. In particular, we first squeeze the global spatial information of each feature into channel-wise statistics using a squeeze operation $\mathcal{F}^{\text{SQ}}$:
\begin{align}\small
\bm{c}_t^{\text{SAL}} &= \mathcal{F}^{\text{SQ}}({\bm{X}}{}_t^{\text{SAL}}) \in\mathbb{R}^C, \\
\bm{c}_t^{\text{INS}} &= \mathcal{F}^{\text{SQ}}({\bm{X}}{}_t^{\text{INS}})\in\mathbb{R}^C,
\end{align}
where $\mathcal{F}^{\text{SQ}}$ is a global average pooling layer. Then, the two channel-wise descriptors are concatenated together to obtain $\bm{c}\in\mathbb{R}^{2C}$, which is then processed by an excitation operation $\mathcal{F}^{\text{EX}}$:
\begin{equation}\small
%\textit{\textbf{c}}' = f_{ex}(\textit{\textbf{c}}; W) = (W_2\delta(W_1\textit{\textbf{c}}))\in\mathbb{R}^{2C}, \\
\bm{z}\!=\!\mathcal{F}^{\text{EX}}(\bm{c}; \bm{W}) \!=\!\text{softmax}(\text{reshape}(\bm{W}_2\delta(\bm{W}_1\bm{c})))\!\in\!\mathbb{R}^{2\!\times\!C},
\end{equation}
where $\bm{W}_{\{1,2\}}$ denotes two fully connected layers, $\delta$ refers to the ReLU function. Note that, after the fully connected layers, we reshape the corresponding vector into $2\!\times\!C$ (which consists of two vectors $\alpha\!\in\!\mathbb{R}^{C}$ and $\beta\!\in\!\mathbb{R}^{C}$). Then, we  apply a \textit{softmax} function to ensure $\alpha+\beta=1$. Finally, we obtain the features $\hat{\bm{X}}{}_t^{\text{SAL}}$ and $\hat{\bm{X}}{}_t^{\text{INS}} $ as follows:
\begin{align}\small
\hat{\bm{X}}{}_t^{\text{SAL}} &= \alpha_1\bm{X}_t^{\text{SAL}} + \beta_1\bm{X}_{t-1}^{\text{SGM}},\\
\hat{\bm{X}}{}_t^{\text{INS}} &= \alpha_2\bm{X}_t^{\text{INS}} + \beta_2\bm{X}_{t-1}^{\text{SGM}},
\end{align}
where $\alpha_{1,2}$, $\beta_{1,2}$ denote the attention vectors to weight contributions of different features.

\noindent\textbf{Salient Object Estimation Head.}
Given $\hat{\bm{X}}{}_t^{\text{SAL}}$, we propose a simple yet effective decoder for salient object estimation. Specifically, we first bilinearly upsample $\hat{\bm{X}}{}_t^{\text{SAL}}$ by a factor of 2 and then concatenate it with the corresponding low-level features from the backbone network with the same spatial resolution (\ie, \texttt{res3}). The upsampled features are further processed by a $5\times5$ convolutional layer, and then upsampled again by a factor of 2. After concatenating then with the features in \texttt{res2}, we process them with two consecutive $5\times5$ convolutional layers and one $1\times1$ convolutional layer to obtain the foreground estimation result $S_t$. Finally, the cross entropy loss is employed to evaluate the result against the corresponding ground-truth.

\noindent\textbf{Instance-Aware Segmentation Head.}
The instance segmentation head has a similar architecture to the foreground estimation head, only differing in that it predicts two outputs, \ie, an object center heatmap and a pixel offset field. Inspired by recent point-aware object representations~\cite{zhou2019objects,lee2020centermask}, we represent each object instance by its center. For dense prediction, we additionally predict the offset of each pixel to its corresponding instance center. During training, ground-truth instance centers are encoded by a 2-D Gaussian with a standard deviation of $10$ pixels. We adopt the mean squared error (MSE) loss to minimize the distance between the predicted heatmaps and Gaussian-encoded ground-truths. For the offset learning, we employ the $l_1$ loss for optimization, which is only activated at pixels belonging to foreground object regions. During inference, predicted foreground pixels are grouped to the closest object center based on the predicted offset field, completing the instance grouping.

\subsection{Target-Aware  Tracking Network (T-Net)}\label{sec:tnet}

\noindent\textbf{Target-Specific Tracking.}
For each object instance, we build a target-specific appearance model to discriminate the target from background distractors. Specifically, we instantiate the model with a two-layer fully convolutional network as in ~\cite{robinson2020learning}:
\begin{equation}\label{eq:1}\small
\bm{S} = \mathcal{T}(\bm{X}; \bm{W}) = \bm{W}_2 * (\bm{W}_1 * \bm{X}),
\end{equation}
where $\bm{X}$ denotes the image feature of frame $I\!\in\!\mathcal{V}$, $\bm{W}\! =\! \{\bm{W}_1, \bm{W}_2\}$ are the network parameters of the two convolutional layers, and $*$ indicates the convolution operator. $\bm{S}$ is the output of the T-Net $\mathcal{T}$, which indicates coarse segmentation score prediction. For the semi-supervised VOS task\!~\cite{robinson2020learning}, Eq.~\eqref{eq:1} is trained over a set of $m$ training samples $\mathcal{S} = \{(\bm{X}_j, \bm{y}_j)\}_{j=1}^m$ collected from the ground-truth annotations in the first frame, by minimizing the following objective function:
\begin{equation}\label{eq:lr}\small
\mathcal{L}(\bm{W}; \mathcal{S})\!=\!\sum_{j}\alpha_j\|\mathcal{T}(\bm{X}_j; \bm{W}) - \bm{y}_j\|^2 + \sum_k\lambda_k\|\bm{W}_k\|^2,
\end{equation}
where $\bm{y}_j$ denotes the target label of $\bm{X}_j$ and $\alpha_j\geq0$ is the weight of $\bm{X}_j$, controlling the impact of the sample on the objective. The parameter $\lambda$ balances the contributions of the objective term as well as the regularization term. Note that the training sample set $\mathcal{S}$ is critical for robust model learning, especially in the unsupervised setting. In contrast to the semi-supervised setting, no ground-truth $\bm{y}_0$ is available for model training in the first video frame, and directly training on the segment proposals generated from D-Net is more prone to drifting. To address this, in our approach, the segment proposal from D-Net for each target serves as the pseudo ground-truth label $\check{\bm{y}}_0$ for initial model learning. Unlike\!~\cite{robinson2020learning}, which regularly updates the training set $\mathcal{S}$ using tracked segments, we design heuristic strategies for the backward verification of tracking results, and adaptively update $\mathcal{S}$. This enables our model to be robust to the noises in $\check{\bm{y}}_0$, and greatly boosts the performance.

\noindent\textbf{Target-Agnostic Verification.}
Let $\check{\bm{y}}_j$ and $\mathcal{Y}$ denote the tracking result of a target in frame $I_j$ and its corresponding tracklet, respectively. We aim to verify the consistency between $\check{\bm{y}}_j$ and $\mathcal{Y}$, as well as find a better possible candidate from the object proposal set. This is achieved by matching the object proposals in the current frame with historical tracking results. To promote the reliablity of verification, we conduct the matching in a target-agnostic manner, using a pre-trained ReID network~\cite{luiten2020unovost}. For each object proposal $p$, its matching score with $\mathcal{Y}$ is computed as:
\begin{equation}\label{eq:score}\small
\!\!\!s(p, \mathcal{Y})\!=\!( \text{cos}(p, \check{\bm{y}}_j) + \text{cos}(p, \check{\bm{y}}_0) ) * \mathbbm{1}(\text{IoU}(p, \check{\bm{y}}_j) > 0.5),
\end{equation}
where $\text{cos}(\cdot,\cdot)$ indicates the cosine similarity between two ReID embeddings, $\text{IoU}(\cdot,\cdot)$ measures the intersection-over-union between two segments, and $\mathbbm{1}(\cdot)\!\in\!\{0,1\}$ is the indicator function. In Eq.~\eqref{eq:score}, we first examine the overlap ratio between $p$ and $\check{\bm{y}}_j$, which is used to truncate the ReID similarities. For more reliable matching,  we compare $p$ with the most recent and the most distant tracking results, \ie, $\check{\bm{y}}_j$ and $\check{\bm{y}}_0$. This allows our model to conduct sequential modeling, while at the same time dealing with long-term semantic consistency, leading to more robust matching results. Based on Eq.~\eqref{eq:score}, we find the proposal with the highest score $s$ with $\mathcal{Y}$. If $s$ is above a threshold $th_{\text{reid}}$ (\eg, 0.6), we replace the current tracking result $\check{\bm{y}}_j$ with the corresponding proposal; otherwise, we keep $\check{\bm{y}}_j$ unchanged. Besides, we discover new targets if all corresponding proposals have zero matching scores with all existing tracklets as well as very small IoUs ($<\!0.1$) with tracking results in the current frame. This provides high flexibility to our model in dealing with occlusions and discovering newly-appearing objects.

\noindent\textbf{Adaptive Memory Updating.}
Once the tracking result $\check{\bm{y}}_j$ is verified, we add a new sample $\{\bm{X}_j, \check{\bm{y}}_j\}$ into the training set $\mathcal{S}$ in order to guide the learning of latest appearance features. The sample is first assigned a weight $\alpha_j\!=\!(1-\gamma)^{-1}\alpha_{j-1}$, where $\alpha_0\!=\!\gamma$. For proposals with $m\!>\!th_{\text{reid}}$, we double the corresponding weight $\alpha_j$ so that the model can put more emphasis on reliable object proposals. Then, we normalize all weights in the training set to unity. During inference, if $m\!>\!th_{\text{reid}}$, we intermediately update the appearance model in the frame; otherwise, we update the model every eight frames.

%we determine to adaptively update the memory using the new sample $\{\mathbf{x}_j, \mathbf{\tilde{y}}_j, \alpha_j\}$. The sample is first given a weight $\alpha_j\!=\!(1-\eta)^{-1}\alpha_{j-1}$, where $\alpha_0\!=\!\eta$. Besides, if $\mathbf{\tilde{s}}\!>\!th_{\text{reid}}$, we double the corresponding weight $\alpha_j$ so that the model can put more emphasis on relible object proposals. All the weights are then normalized to unity. During inference, if $\mathbf{\tilde{s}}\!>\!th_{\text{reid}}$, we intermediately update the appearance model at the frame; otherwise, we update the model every 8 frames.

%\noindent\textbf{Segmentation Network.}
%The appearance model  produces a coarse segmentation output $\mathbf{u}\!=\!D(\mathbf{x};\mathbf{w})$. It is then passed to a segmentation network $S$ to obtain a high-resolution segmentation. Our segmentation network consists of two modules: 1) a target segmentation encoder\!~\cite{robinson2020learning} that merges the segmentation scores with backbone features; and 2) a boundary-aware refinement module\!~\cite{zhou2020motion} to produce accurate segmentation with crisp boundaries. %More details can be found in the corresponding papers.

\begin{figure*}[t]

	\begin{minipage}[t]{\textwidth}
		\centering
		\begin{threeparttable}
			\resizebox{1\textwidth}{!}{
				\setlength\tabcolsep{5pt}
				\renewcommand\arraystretch{1}
				\begin{tabular}{c|r||c|c|ccc|ccc}
					\hline\thickhline
					\rowcolor{mygray}
					Dataset & Method & Pub. &  $\mathcal{J}\!\&\mathcal{F}$ Mean$\uparrow$ & $\mathcal{J}$ Mean$\uparrow$ & $\mathcal{J}$ Recall$\uparrow$ & $\mathcal{J}$ Decay$\downarrow$ & $\mathcal{F}$ Mean$\uparrow$ & $\mathcal{F}$ Recall$\uparrow$ & $\mathcal{F}$ Decay$\downarrow$ \\ \hline \hline
					\multirow{9}{*}{\textit{val}} & RVOS\!~\cite{ventura2019rvos} & CVPR$_{19}$ & 41.2 & 36.8	& 40.2 & 0.5 & 45.7 & 46.4 & \textit{\color{blue}1.7}\\
					& $^\ddag$OF-Tracker\!~\cite{athar2020stem} & - & 54.6 & 53.4& 60.9& -1.3 &55.9& 63.0 &1.1 \\
					& $^\ddag$RI-Tracker\!~\cite{athar2020stem} & - & 56.9 & 55.5& 63.3& 2.7& 58.2& 64.4& 6.4 \\

					& PDB\!~\cite{song2018pyramid}    & ECCV$_{18}$ & 55.1 & 53.2	& 58.9 & 4.9 & 57.0 & 60.2 & 6.8\\
					& AGS\!~\cite{Wang2020PayingAT} & CVPR$_{19}$ & 57.5 & 55.5 & 61.6 & 7.0 & 59.5 & 62.8 & 9.0 \\ 
					& ALBA\!~\cite{gowda2020alba} & BMVC$_{20}$ &  58.4 & 56.6 & 63.4 & 7.7 &60.2 & 63.1 & 7.9 \\ 
					& MATNet\!~\cite{zhou2020matnet} & TIP$_{20}$ & 58.6 & 56.7 & 65.2 & -3.6 & 60.4 & 68.2 & 1.8 \\ 
					& AGNN\!~\cite{wang2019zero} & ICCV$_{19}$ & 61.1 & 58.9 & 65.7 & 11.7 & 63.2 & 67.1 & 14.3 \\
					& STEm-Seg\!~\cite{athar2020stem} & ECCV$_{20}$ & 64.7 & 61.5 & 70.4 & \textbf{{-4.0}} & 67.8 & 75.5 & 1.2 \\
					& $^*$UnOVOST\!~\cite{luiten2020unovost} & WACV$_{20}$ & \textbf{67.9} & \textbf{66.4} & \textbf{76.4} & -0.2 & \textbf{69.3} & \textbf{76.9} & \textbf{0.0} \\ \cline{2-10} 
					& \textbf{Ours} & - & 65.0 & 63.7 & 71.9 & 6.9 & 66.2 & 73.1 & 9.4  \\
					\hline
				\end{tabular}
			}
		\end{threeparttable}
	\end{minipage}
	
	\begin{minipage}[t]{\textwidth}
		\vspace{8pt}
		\centering
		\begin{threeparttable}
			\resizebox{1\textwidth}{!}{
				\setlength\tabcolsep{5pt}
				\renewcommand\arraystretch{1}
				\begin{tabular}{c|r||c|c|ccc|ccc}
					\hline\thickhline				\rowcolor{mygray}
					Dataset & Method & Pub. &  $\mathcal{J}\!\&\mathcal{F}$ Mean$\uparrow$ & $\mathcal{J}$ Mean$\uparrow$ & $\mathcal{J}$ Recall$\uparrow$ & $\mathcal{J}$ Decay$\downarrow$ & $\mathcal{F}$ Mean$\uparrow$ & $\mathcal{F}$ Recall$\uparrow$ & $\mathcal{F}$ Decay$\downarrow$ \\ \hline \hline
					\multirow{8}{*}{\textit{test-dev}}& RVOS\!~\cite{ventura2019rvos} & CVPR$_{19}$ & 22.5	& 17.7	& 16.2 & 1.6	 & 27.3 &	24.8 & 1.8 \\
					& PDB\!~\cite{song2018pyramid}  & ECCV$_{18}$ & 40.4	& 37.7	& 42.6 & 4.0 & 43.0 & 44.6 & 3.7\\
					%& MuG-W\!~\cite{lu2020learning} & CVPR$_{20}$ & 41.7 & 38.9 & 44.3 &  \textbf{-2.7}& 44.5 & 46.6 & \textbf{-1.7}\\
					& AGS\!~\cite{Wang2020PayingAT} & CVPR$_{19}$ & 45.6	& 42.1 & 48.5 & 2.6 & 49.0 & 51.5 & 2.6  \\
					\cline{2-10}
					%& Oxf-CAS\!~\cite{DAVIS2019-Unsupervised-2nd} & DAVIS$_{19}$ & 56.5 & 51.7 & 59.9 & 21.7 & 61.4 & 65.7 & 15.7 \\
					& MSP\!~\cite{DAVIS2020-Unsupervised-1st} & DAVIS$_{20}$ & 57.9 & 52.9 & 60.4 & 16.7 & 63.0 & \textbf{69.5} & 20.5 \\
					& $^*$UnOVOST\!~\cite{luiten2020unovost} & DAVIS$_{19}$ & 58.0 & 54.0	& 62.9 &	3.5 &	62.0 &	66.6 &	6.6  \\
					\cline{2-10}
					
					& \textbf{Ours} & - & \textbf{59.8}  & \textbf{56.0} & \textbf{65.1} & 7.8  & \textbf{63.7} & 68.4 &  11.0 \\
					\hline
				\end{tabular}
			}
		\end{threeparttable}
	\end{minipage}
	\vspace{-8pt}
	\makeatletter\def\@captype{table}\makeatother\caption{\small\textbf{Quantitative video multi-object segmentation results on the \texttt{val} and \texttt{test-dev} sets of DAVIS$_{17}$} in terms of region similarity $\mathcal{J}$ and boundary accuracy $\mathcal{F}$. `DAVIS$_{19}$' and `DAVIS$_{20}$' indicate the unsupervised tracks of the DAVIS 2019 and 2020 challenges, respectively. $^\ddag$: baseline methods implemented in \!~\cite{athar2020stem}. $^*$: methods has complex heuristic post-processing.}
	\label{table:davis}
	\vspace{-8pt}
\end{figure*}

%\section{Detailed Network Architecture}

\section{Experiment}

In this section, we present the experimental results of our approach. We first elaborate on the datasets, training and testing settings in~\S\ref{sec:setting}. Then, we investigate the performance of our method for the unsupervised video multi-object segmentation task in \S\ref{sec:davis} and video instance segmentation task in \S\ref{sec:vis}, respectively. Visual comparison results are presented in \S\ref{sec:visual}. We conduct detailed ablative experiments in~\S\ref{sec:ablation}. Finally, we provide run time analysis to quantify the efficiency of the system in~\S\ref{sec:runtime}. 

\subsection{Experimental Settings}\label{sec:setting}
\noindent\textbf{Datasets.} We conduct experiments on two popular datasets:
\begin{itemize}[leftmargin=*]
	\setlength{\itemsep}{0pt}
	\setlength{\parsep}{-2pt}
	\setlength{\parskip}{-0pt}
	\setlength{\leftmargin}{-8pt}
	\vspace{-4pt}%In particular, semantic
	\item \textbf{DAVIS$_{17}$}~\cite{caelles20192019} for \textit{video multi-object segmentation}. The dataset consists of $120$ high-quality videos in total.  These videos are further split into $60$ for \texttt{train}, $30$ for \texttt{val} and 
	$30$ for \texttt{test-dev}. In our experiments, we train our models only on the \texttt{train} split, without any additional data. Then, we evaluate the performance of our approach on \texttt{val} and \texttt{test-dev}.    
	
	\item \textbf{YouTube-VIS}~\cite{xu2018youtube} for \textit{video instance segmentation}. It contains $2,\!883$ high-resolution videos collected from YouTube, covering more than 131K object instances. Different from DAVIS$_{17}$ in which objects are category-agnostic, objects in YouTube-VIS are labeled with one semantic category out of 40 categories. Therefore, the task in YouTube-VIS not only requires the algorithms to segment consistent objects but also assign each object a category label. We use this dataset to examine the performance of our model in more challenging scenarios.
	
	\vspace{-5pt}
\end{itemize}

\noindent\textbf{Training Phase.} We train the D-Net on the training set of DAVIS$_{17}$ and YouTube-VIS. During training, each sample is randomly augmented with a scaling factor of [0.8, 1.5] and horizontal flipping, and is then cropped to $640\!\times\!640$. For optimization, we use the standard SGD solver, with a momentum of 0.9 and weight decay of 5e-4. We utilize the polynomial annealing procedure to schedule the learning rate. For the T-Net, we use ResNet-101 as the backbone network of the appearance model. The layer $\textit{\textbf{w}}_1$ is a $1\times1$ convolutional layer that reduces the channel of input features to 96 while $\textit{\textbf{w}}_2$ is a $3\times3$ convolutional layer with one output channel. The two layers are optimized online using the Gauss-Newton algorithm\!~\cite{robinson2020learning} with the default settings, which leads to significantly faster convergence than other gradient descent-based approaches. 

\noindent\textbf{Testing Phase.} Given a test video, we run our instance discrimination network and target-aware adaptive tracking network to process each frame sequentially. The input image sizes for the two networks are separately set to $480\times 854$ and $473\times 473$.  Our model requires no additional post-processing components (\eg, CRF), which guarantees high efficiency (10 FPS) against state of the arts. 

\noindent\textbf{Evaluation Metrics.} 
We follow the standard evaluation settings used in each dataset for evaluation. 1) For DAVIS$_{17}$, we report the performance in terms of region similarity $\mathcal{J}$, boundary accuracy $\mathcal{F}$, and the overall metric $\mathcal{J}\&\mathcal{F}$. The evaluation scores on the \texttt{test-dev} set are obtained from the evaluation server of the DAVIS$_{20}$ challenge, since the ground-truths of the set are private. 2) For YouTube-VIS, we follow~\cite{xu2018youtube} to use average precision (AP) and average recall (AR) as the metrics, which are adapted from the image instance segmentation task to the video instance segmentation task.
\begin{table}[t]
	\centering	
	
	\resizebox{0.49\textwidth}{!}{
		\setlength\tabcolsep{6pt}
		\renewcommand\arraystretch{1.05}
		
		\begin{tabular}{r||ccccc}
			\hline\thickhline				\rowcolor{mygray}
			Method & mAP & AP$_{50}$ & AP$_{75}$  & AR$_{1}$ & AR$_{10}$ \\ \hline\hline
			
			$^\ddag$DeepSORT\!~\cite{wojke2017simple} & 26.1 & 42.9 & 26.1 & 27.8 & 31.3 \\
			$^\ddag$FEELVOS\!~\cite{voigtlaender2019feelvos} & 26.9 & 42.0 & 29.7 & 29.9 & 33.4 \\
			$^\ddag$OSMN\!~\cite{yang2018efficient} & 27.5 & 45.1 & 29.1 & 28.6 & 33.1 \\	
			MaskTrack R-CNN\!~\cite{yang2019video}  &30.3 &51.1 &32.6& 31.0& 35.5\\
			SeqTracker\!~\cite{yang2019video} & 27.5 & 45.7 & 28.7 & 29.7 & 32.5 \\
			STEm-Seg\!~\cite{athar2020stem} & 35.0 & 56.0 & 38.6 & 34.4 & 41.7 \\ \hline
			\textbf{Ours} & \textbf{37.1} & \textbf{57.1} & \textbf{40.9} & \textbf{34.8} & \textbf{43.2} \\
			\hline
	\end{tabular}}
	\caption{\small \textbf{Quantitative video instance segmentation results on Youtube-VIS \texttt{val}}, in terms of AP and AR. The baselines denoted with $^\ddag$ were implemented by the authors in\!~\cite{yang2019video}.}
	\label{table:ytb}
	\vspace{-10pt}
\end{table}

\begin{figure*}[t]
	\begin{center}
		\includegraphics[width=\linewidth]{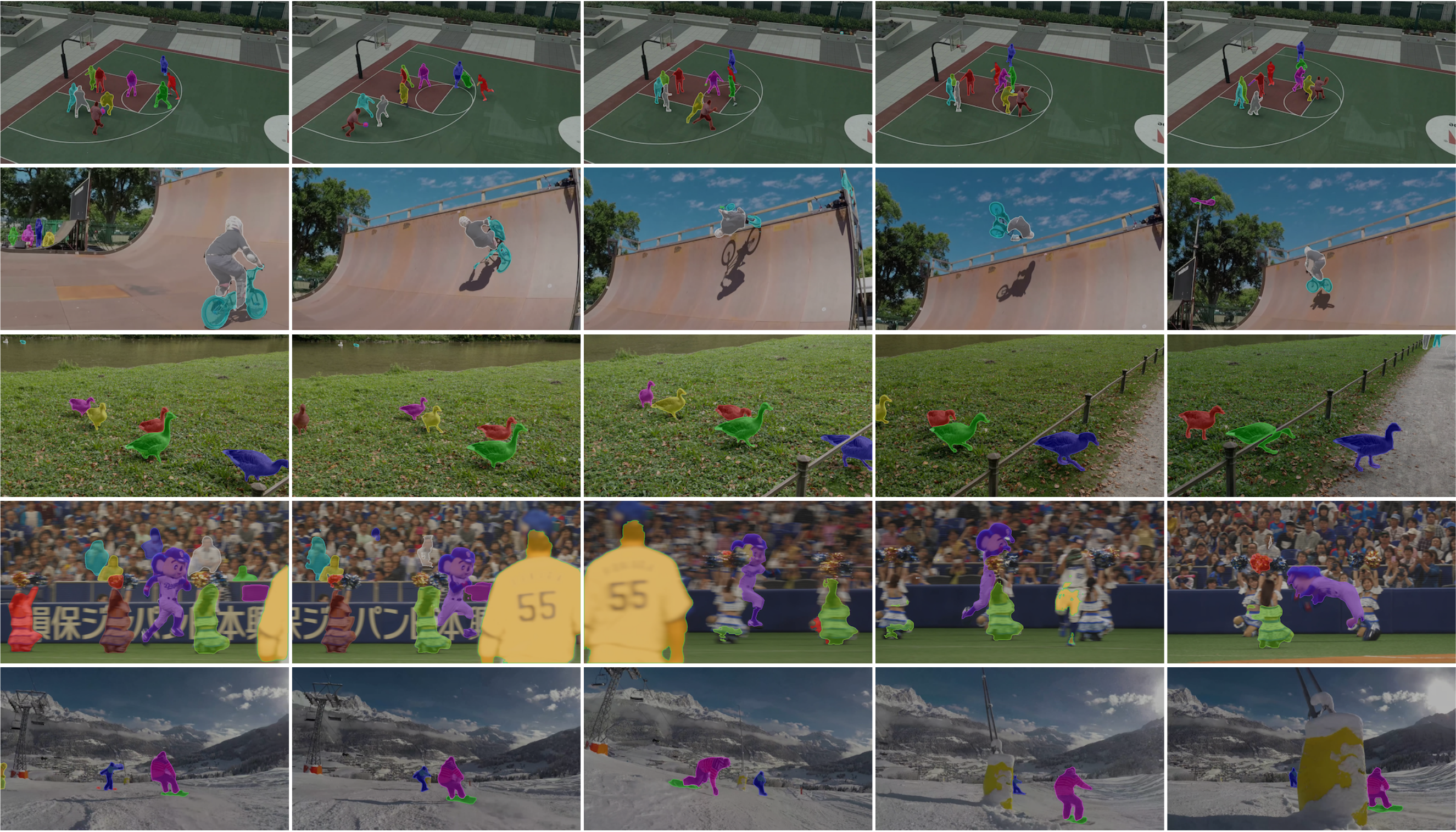}
	\end{center}
	\vspace{-16pt}
	\captionsetup{font=small}
	\caption{\small Qualitative results of multi-object segmentation masks on DAVIS$_{17}$ \texttt{test-dev}. From top to bottom: \textit{basketball-game}, \textit{bmx-rider}, \textit{ducks}, \textit{mascot}, and \textit{snowboard-race}. }
	\label{fig:5}
	\vspace{-10pt}
\end{figure*}

%\begin{figure*}[t]
%\begin{center}
%	\includegraphics[width=\linewidth]{figure/Fig4.pdf}
%\end{center}
%%\captionsetup{font=small}
%\caption{Visual results of multi-object segmentation masks on DAVIS$_{17}$ \texttt{val}. From top to bottom: \textit{dance-twirl}, \textit{dogs-jump}, \textit{gold-fish}, \textit{india}, and \textit{lab-coat}. }
%\label{fig:4}
%\end{figure*}

\subsection{Performance Comparison on DAVIS$_{17}$}\label{sec:davis}
We compare our approach with state-of-the-art video multi-object segmentation methods on the DAVIS$_{17}$ benchmark. In addition to recently published works (\eg, RVOS\!~\cite{ventura2019rvos}, PDB\!~\cite{song2018pyramid}, AGS\!~\cite{Wang2020PayingAT} , ALBA\!~\cite{gowda2020alba}, MATNet\!~\cite{zhou2020matnet}), we also include some top-ranked solutions (\ie, UnOVOST\!~\cite{luiten2020unovost}, MSP\!~\cite{DAVIS2020-Unsupervised-1st}) from the unsupervised tracks of the DAVIS-2019 and DAVIS-2020 VOS challenges. This leads to a more comprehensive examination of the proposed approach. As reported in Table~\ref{table:davis}, on DAVIS$_{17}$ \texttt{val}, our approach achieves the second-best overall results across most metrics. It is slightly worse than UnOVOST, the champion solution in DAVIS-2019 VOS challenge. However, we emphasize that UnOVOST is computationally expensive, requiring not only Mask R-CNN for instance proposal generation, but also needing to compute optical flow for motion estimation. Complex post-processing and heuristics also make the method unsuitable for many practical applications. In addition, we see that our approach outperforms all other comparative approaches. 

On  DAVIS$_{17}$ \texttt{test-dev}, our method outperforms all competitors, including all top solutions in the challenges. Keeping in mind that \texttt{test-dev} is more challenging than \texttt{val} and the ground-truths are kept private, the good performance over this set can better support our approach.

\begin{table}[t]
	\centering
	\setlength{\tabcolsep}{0.3em}	
	
	\resizebox{0.49\textwidth}{!}{
		\setlength\tabcolsep{4pt}
		\renewcommand\arraystretch{1.05}
		\begin{tabular}{r||ccc|ccc}
			\hline\thickhline				\rowcolor{mygray}
			Model & AP & AP$_{50}$ & AP$_{75}$ & AP$_{S}$ & AP$_{M}$ & AP$_{L}$ \\ \hline\hline
			Mask R-CNN\!~\cite{he2017mask} & 48.1 & 70.0 & 51.1 & 29.9 & 51.8 &  62.2 \\
			SOLO\!~\cite{wang2019solo} & 46.8 & 68.6 & 49.3 & 28.7 & 50.4 & 60.9  \\ \hline
			D-Net \textit{w/o} SGM  & 50.3 & 71.9 & 53.3 & 28.9 & 53.6 & 65.3  \\
			\textbf{Full D-Net} & \textbf{52.0} & \textbf{73.3} & \textbf{54.9} & \textbf{30.3} & \textbf{55.4} & \textbf{67.5} \\
			\hline
	\end{tabular}}
	\caption{\small\textbf{Ablation study of D-Net on DAVIS$_{17}$ \texttt{val}.} We report the AP scores with and without the segmentation guidance module (\ie, SGM). For comparison, we report the performance of Mask R-CNN\!~\cite{he2017mask} and SOLO\!~\cite{wang2019solo}, which are representative methods for top-down and bottom-up instance segmentation models, respectively. All the models use ResNet-50 as the backbone. See \S\ref{sec:ablation}.}
	\label{table:instanceseg}
	\vspace{-5pt}
\end{table}

\subsection{Performance Comparison on YouTube-VIS}\label{sec:vis}
We further examine the performance of the proposed approach on YouTube-VIS, which requires not only segmenting and tracking objects, but also recognizing their semantic categories. To this end, we modify the foreground estimation head in the D-Net to predict semantic labels of each pixel (instead of original binary labels) following general semantic segmentation networks~\cite{long2015fully,chen2017deeplab}. We train all the networks on the training data of YouTube-VIS. As reported in Table~\ref{table:ytb}, our approach outperforms all the comparative methods with respect to all metrics. We improve the AP by \textbf{+2.1\%} in comparison with the most recent model STEm-Seg~\cite{athar2020stem}. Our approach also significantly outperforms existing two-stage methods, like MaskTrack R-CNN~\cite{yang2019video}, demonstrating its superiority.

\begin{table}[t]
	\centering
	
	\resizebox{0.49\textwidth}{!}{
		\setlength\tabcolsep{1pt}
		\renewcommand\arraystretch{1.05}
		\begin{tabular}{r||c|cc|cc}
			\hline\thickhline
			\rowcolor{mygray}
			Variant & $\mathcal{J}\!\&\!\mathcal{F}$ Mean & $\mathcal{J}$ Mean & $\mathcal{J}$ Recall & $\mathcal{F}$ Mean & $\mathcal{F}$ Recall \\ \hline\hline
			\textit{w/o.} target verification & 61.3 & 58.2 & 63.5 & 61.9 & 66.3 \\
			\textit{w/o.} memory updating & 61.2 & 59.4 & 66.3 & 62.8 & 69.2 \\
			\hline
			\textbf{Full Model} & \textbf{65.0} & \textbf{63.7} & \textbf{71.9} & \textbf{66.2} & \textbf{73.1} \\ \hline
	\end{tabular}}
	%	\captionsetup{font=small}
	\caption{\small\textbf{Key component analysis of the proposed T-Net on DAVIS$_{17}$ \texttt{val}.} See \S\ref{sec:ablation} for details.}
	\label{table:ablation}
	\vspace{-5pt}
\end{table}

\subsection{Qualitative Result}\label{sec:visual}
In \figref{fig:5}, we show the qualitative segmentation results of our approach on \texttt{test-dev}. Different colors are used to indicate different object instances. From the figures, we can see the remarkable performance of the proposed approach in 1) accurately discovering distinct objects in complex scenarios (\eg, low-light illumination in \textit{gold-fish}), as well as 2) producing robust and temporally coherent object tracking across the sequence. Moreover, our approach shows good performance in dealing with various challenging factors, such as, occlusions, scale variations, fast motion.

\subsection{Diagnostic Experiment}\label{sec:ablation}
%In this section, we examine the contributions of each essential component in our approach.

\noindent\textbf{Segmentation Guidance Module.} To demonstrate the superiority of the D-Net in comparison with other counterparts, we compare it with two baseline methods (\ie, Mask R-CNN~\cite{he2017mask} and SOLO~\cite{wang2019solo}) in terms of category-agnostic instance segmentation on DAVIS$_{17}$ \texttt{val}. We also examine the performance of the D-Net with and without the SGM. For fair comparison, we  follow the standard setting~\cite{he2017mask,wang2019solo} to use mAP as the metric for evaluation. As reported in  Table~\ref{table:instanceseg}, the D-Net \textit{w/o} SGM achieves obvious performance improvement against the two baselines ($+2.2\%$ in terms of AP). By incorporating the SGM, our full model further improves the AP by $\textbf{+1.7\%}$, thereby demonstrating the effectiveness of the SGM module.

\noindent\textbf{Key Components in T-Net.} We further conduct experiments to verify the essential components (\ie, target verification and memory updating modules) in T-Net. We examine the performance by discarding each module once at a time. As summarized in Table~\ref{table:ablation}, the performance drops significantly after removing each module compared with the full model, proving their efficacy.

\begin{table}[t]
	\centering
	
	\resizebox{0.49\textwidth}{!}{
		\setlength\tabcolsep{4pt}
		\renewcommand\arraystretch{1.01}
		\begin{tabular}{r||ccc|c}
			\hline\thickhline\rowcolor{mygray}
		 & Instance & Foreground & Tracking / & \\ 
			\rowcolor{mygray}
			\multirow{-2}{*}{Method}	& Proposal & Estimation &  Matching &  \multirow{-2}{*}{Total Time (s)}\\
			\hline\hline
			RVOS\!~\cite{ventura2019rvos} & - & - & 0.07  &  0.07 \\
			PDB\!~\cite{song2018pyramid} & 0.74 &  0.70 & 0.03 & 1.47  \\
			AGS\!~\cite{Wang2020PayingAT} & 0.74 & 0.10 & 0.03 & 0.87 \\
			MATNet\!~\cite{zhou2020matnet} & 0.74 & 0.75 & 0.03 & 1.52\\
			UnOVOST\!~\cite{luiten2020unovost} & 0.74 & 0.20 & 0.08 & 1.02 \\ \hline
			Ours & 0.05 & - & 0.06 & 0.11 \\
			\hline
	\end{tabular}}
	
	\caption{\small\textbf{Runtime analysis} (second/frame) on DAVIS$_{17}$ \texttt{val}. Note that our approach is much faster than existing two-stage methods. Although slightly slower than RVOS, our approach has a better tradeoff between segmentation accuracy and efficiency.}
	\label{table:runtime}
	\vspace{-10pt}
\end{table}

\subsection{Runtime Comparison}\label{sec:runtime}

\begin{figure}[t]
	\begin{center}
		\includegraphics[width=0.8\linewidth]{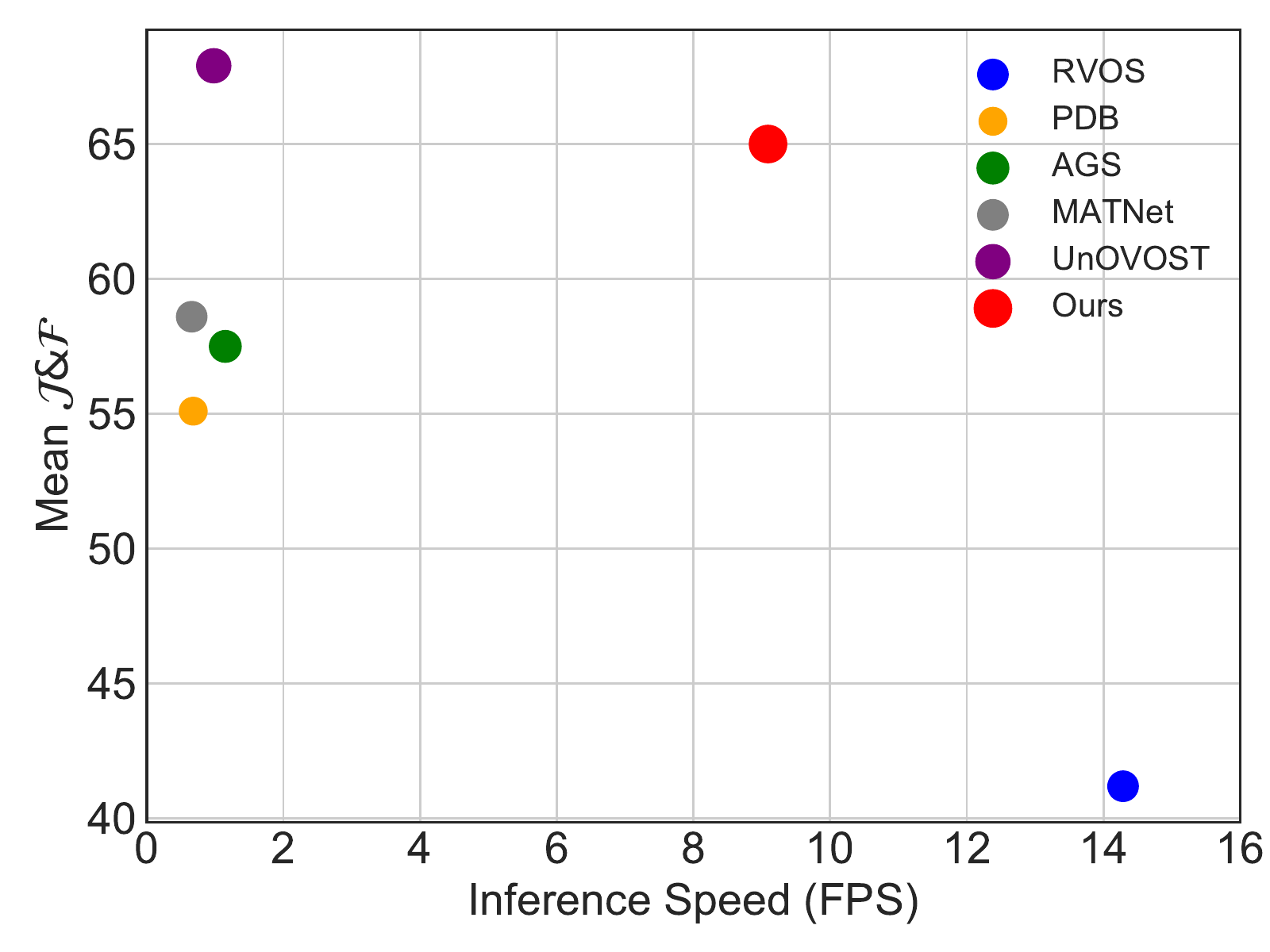}
	\end{center}
	\vspace{-16pt}
	\captionsetup{font=small}
	\caption{\small Trade-off between inference speed (\textit{x}-axis) and segmentation accuracy (\textit{y}-axis) on DAVIS$_{17}$ \texttt{val}. Our approach demonstrates compelling performance with high efficiency.}
	\label{fig:6}
	\vspace{-10pt}
\end{figure}

In addition to segmentation accuracy, runtime efficiency is also an important dimension for evaluating the usability of U-VOS algorithms. For this reason, we conduct a runtime analysis on DAVIS$_{17}$ \texttt{val} for a more comprehensive comparison. Five representative methods are used for comparison, including RVOS\!~\cite{ventura2019rvos}, PDB\!~\cite{song2018pyramid}, AGS\!~\cite{Wang2020PayingAT}, MATNet\!~\cite{zhou2020matnet}, and UnOVOST\!~\cite{luiten2020unovost}.  For each model, we report the inference speeds in terms of three components, \ie, instance proposal, foreground estimation (or salient object estimation), and instance tracking (or matching). Note that most comparative methods simply claim to use MASK R-CNN for instance proposal generation without revealing too many details (\eg, backbones). Thus, we directly use the value (0.74 s) reported in UnOVOST\!~\cite{luiten2020unovost} for all the methods as reference. The analysis results are summarized in Table~\ref{table:runtime}. We observe that, since our approach formulates instance proposal and foreground estimation in a unified framework, it requires much less time to generate instance proposals. Further, with the efficient target-aware tracking network, our approach can run at about 10 FPS, taking 0.11 s to process one image with 480p resolution. Though it is slightly slower than RVOS (0.07 s), we have seen from Table~\ref{table:davis} that our approach is able to produce considerably more accurate segmentation results. 
~\figref{fig:6} depicts a visualization of the trade-off between accuracy and efficiency of representative algorithms on the validation set of DAVIS$_{17}$. As can be seen, our approach achieves the best trade-off.

%\begin{table}[t]
%	\centering
%	\caption{Runtime comparison}
%	\resizebox{0.49\textwidth}{!}{
%		\setlength\tabcolsep{2pt}
%		\renewcommand\arraystretch{1.1}
%		\begin{tabular}{r||ccccc|c}
%			\hline\thickhline
%			Time (s)& RVOS & PDB & AGS & MATNet & UnOVOST & Ours\\ \hline\hline
%			Instance Proposal  & - & 0.74 & 0.74 & 0.74 &  0.74 & 0.05 \\
%			Foreground Estimation   & - & 0.70 & 0.10 & 0.75 & 0.20 & - \\
%			Tracking/Matching   & 0.07 & 0.03 & 0.03 & 0.03 & 0.08 & 0.06 \\ \hline
%			Total   & \textbf{\color{red}0.07}& 1.47 & 0.87 & 1.52 & 1.02 & \color{blue}0.11 \\ 
%			\hline
%	\end{tabular}}
%	\label{table:runtime}
%\end{table}

\section{Conclusion}

Unsupervised video object segmentation is significant in empowering machines to automatically understand dynamic real-world scenarios. In this paper, we present a novel approach for multi-object segmentation in unconstrained videos. First, we propose an instance discrimination network to discover salient instance segments in a bottom-up manner. By introducing previously well-segmented masks as guidance for segmenting later frames, the network is able to produce accurate and temporally coherent segments. Second, based on the instance proposals, we design a target-aware adaptive tracking framework to associate the proposals of the same identity across the sequence. By building a target-aware appearance model for each object, our model achieves more robust matching than previous ReID-based methods. Third, we have conducted extensive experiments on two popular benchmarks, \ie, DAVIS$_{17}$ and YouTube-VIS, and the results demonstrate that our approach achieves higher segmentation accuracy against state-of-the-art methods, while running at a faster inference speed.

%{\small\noindent\textbf{Acknowledgment} This work was supported in part by the Beijing Natural Science Foundation under Grant L191004 and L202002.}

{\small
\bibliographystyle{ieee_fullname}
\bibliography{egbib}
}

\end{document}